\documentclass[a4paper]{article}

\usepackage{INTERSPEECH2016}

\usepackage{graphicx}
\usepackage{amssymb,amsmath,bm}
\usepackage{textcomp}
\usepackage{amsfonts}
\usepackage{booktabs}
\usepackage{siunitx}
\usepackage{subcaption}
\usepackage{afterpage}
\usepackage{hyperref}
\usepackage [english]{babel}
\usepackage [autostyle, english = american]{csquotes}
\MakeOuterQuote{"}

\usepackage{array}
\usepackage{arydshln}
\newcommand{\etal}{\textit{et al.}}

\ninept

\title{Learning Visual Reasoning Without Strong Priors}

\makeatletter
\def\name#1{\gdef\@name{#1\\}}
\makeatother
\name{{\em Ethan~Perez$^1$$^2$, Harm~de~Vries$^1$, Florian~Strub$^{3}$,}\\ 
     {\em Vincent~Dumoulin$^1$, Aaron~Courville$^1$$^4$}}

\address{$^1$MILA, Universit\'e of Montr\'eal, Canada;
  $^2$Rice University, U.S.A. \\
  $^3$Univ. Lille, CNRS, Centrale Lille, Inria, UMR 9189 CRIStAL France \\
  $^4$CIFAR Fellow, Canada \\
  {\small \tt ethanperez@rice.edu, mail@harmdevries.com, florian.strub@inria.fr}\\
  {\small \tt dumouliv@iro.umontreal.ca, courvila@iro.umontreal.ca}\\
}

\begin{document}

  \maketitle

  \begin{abstract}
	Achieving artificial visual reasoning --- the ability to answer image-related questions which require a multi-step, high-level process --- is an important step towards artificial general intelligence. This multi-modal task requires learning a question-dependent, structured reasoning process over images from language. Standard deep learning approaches tend to exploit biases in the data rather than learn this underlying structure, while leading methods learn to visually reason successfully but are hand-crafted for reasoning. We show that a general-purpose, Conditional Batch Normalization approach achieves state-of-the-art results on the CLEVR Visual Reasoning benchmark with a 2.4\% error rate. We outperform the next best end-to-end method (4.5\%) and even methods that use extra supervision (3.1\%). We probe our model to shed light on how it reasons, showing it has learned a question-dependent, multi-step process. Previous work has operated under the assumption that visual reasoning calls for a specialized architecture, but we show that a general architecture with proper conditioning can learn to visually reason effectively.
\end{abstract}

\vspace{1em}
\noindent{\bf Note}: \textit{A full paper extending this study is available at \url{http://arxiv.org/abs/1709.07871}, with additional references, experiments, and analysis.}

\section{Introduction} \label{introduction}
    The ability to use language to reason about every-day visual input is a fundamental building block of human intelligence. Achieving this capacity to visually reason is thus a meaningful step towards artificial agents that truly understand the world. Advances in both image-based learning and language-based learning using deep neural networks have made huge strides in difficult tasks such as object recognition \cite{NIPS2012_4824,DBLP:journals/corr/HeZRS15} and machine translation \cite{DBLP:journals/corr/ChoMGBSB14,DBLP:journals/corr/SutskeverVL14}. These advances have in turn fueled research on the intersection of visual and linguistic learning \cite{malinowski2014multi,geman2015visual,antol2015,hvries2016,DBLP:journals/corr/JohnsonHMFZG16}.
    
\begin{figure}[t]
	\centering
    \begin{subfigure}[t]{.2\textwidth}
        \centering
        \includegraphics[height=0.8in]{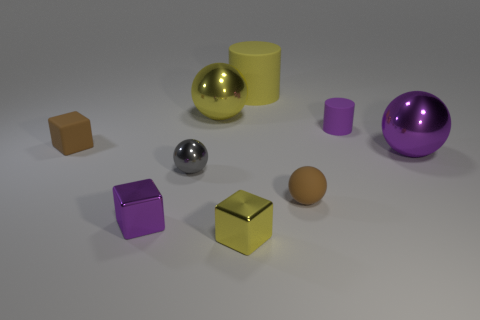}
        \caption{\it{What number of cylinders are small purple things or yellow rubber things?\\
        \bf{Predicted: 2}}}
    \end{subfigure}
    ~ 
    \begin{subfigure}[t]{.2\textwidth}
		\centering
        \includegraphics[height=0.8in]{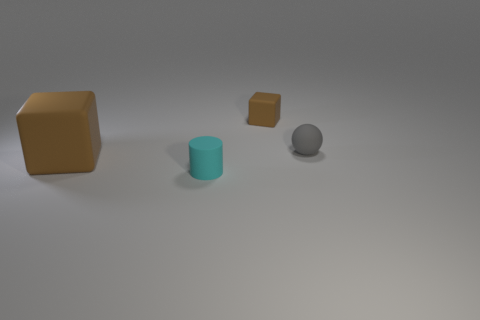}
        \caption{\it{What color is the other object that is the same shape as the large brown matte thing?\\
        \bf{Predicted: Brown}}}
    \end{subfigure}
    \caption{Examples from CLEVR and our model's answer.}
    \label{fig:CLEVR}
    \vskip -1em
\end{figure}

    To this end, \cite{DBLP:journals/corr/JohnsonHMFZG16} recently proposed the CLEVR dataset to test multi-step reasoning from language about images, as traditional visual question-answering datasets such as~\cite{malinowski2014multi,antol2015} ask simpler questions on images that can often be answered in a single glance. Examples from CLEVR are shown in Figure~\ref{fig:CLEVR}. Structured, multi-step reasoning is quite difficult for standard deep learning approaches~\cite{DBLP:journals/corr/JohnsonHMHLZG17,DBLP:journals/corr/SantoroRBMPBL17}, including those successful on traditional visual question answering datasets. Previous work highlights that standard deep learning approaches tend to exploit biases in the data rather than reason~\cite{DBLP:journals/corr/JohnsonHMFZG16,goyal2016making}. To overcome this, recent efforts have built new learning architectures that explicitly model reasoning or relational associations~\cite{DBLP:journals/corr/JohnsonHMHLZG17,DBLP:journals/corr/SantoroRBMPBL17,DBLP:journals/corr/HuARDS17}, some of which even outperform humans~\cite{DBLP:journals/corr/JohnsonHMHLZG17,DBLP:journals/corr/SantoroRBMPBL17}.

    In this paper, we show that a general model can achieve strong visual reasoning from language. We use Conditional Batch Normalization \cite{DBLP:journals/corr/DumoulinSK16,modulating_vision,DBLP:journals/corr/GhiasiLKDS17} with a Recurrent Neural Network (RNN) and a Convolutional Neural Network (CNN) to show that deep learning architectures built without strong priors can learn underlying structure behind visual reasoning, directly from language and images. We demonstrate this by achieving state-of-the-art visual reasoning on CLEVR and finding structured patterns while exploring the internals of our model.

\section{Method} \label{method}

	Our model processes the multi-modal question-image input using a RNN and CNN combined via Conditional Batch Normalization (CBN). CBN has proven highly effective for image stylization \cite{DBLP:journals/corr/DumoulinSK16,DBLP:journals/corr/GhiasiLKDS17}, speech recognition \cite{DynamicLayerNorm}, and traditional visual question answering tasks \cite{modulating_vision}. We start by explaining CBN in Section~\ref{cbn} and then describe our model in Section~\ref{model}.

	\subsection{Conditional batch normalization}
		\label{cbn}

	Batch normalization (BN) is a widely used technique to improve neural network training by normalizing activations throughout the network with respect to each mini-batch. BN has been shown to accelerate training and improve generalization by reducing covariate shift throughout the network~\cite{DBLP:journals/corr/IoffeS15}. To explain BN, we define $\mathcal{B} = \{\bm{F}_{i,.,.,.}\}_{i=1}^N$ as a mini-batch of $N$ samples, where $\bm{F}$ corresponds to input feature maps whose subscripts $c,h,w$ refers to the $c^{th}$ feature map at the spatial location $(h,w)$. We also define $\gamma_c$ and $\beta_c$ as per-channel, trainable scalars and $\epsilon$ as a constant damping factor for numerical stability. BN is defined at training time as follows:
\begin{equation}
BN(\bm{F}_{i,c,h,w} | \gamma_c, \beta_c) = \gamma_c\frac{\bm{F}_{i, c, w, h} - \mathbb{E}_{\mathcal{B}}[\bm{F}_{\cdot, c, \cdot, \cdot}]}{\sqrt{\text{Var}_{\mathcal{B}}[\bm{F}_{\cdot, c, \cdot, \cdot}]+\epsilon}} + \beta_c.
\end{equation}

Conditional Batch Normalization (CBN)~\cite{DBLP:journals/corr/DumoulinSK16,modulating_vision,DBLP:journals/corr/GhiasiLKDS17} instead learns to output new BN parameters $\hat{\gamma}_{i,c}$ and $\hat{\beta}_{i,c}$ as a function of some input $\bm{x_i}$:
    \begin{align}
		\hat{\gamma}_{i,c} = f_c(\bm{x}_i) \qquad ~ \qquad 
		\hat{\beta}_{i,c} = h_c(\bm{x}_i),
	\end{align}
where $f$ and $h$ are arbitrary functions such as neural networks. Thus, $f$ and $h$ can learn to control the distribution of CNN activations based on $\bm{x_i}$.

	Combined with ReLU non-linearities, CBN empowers a conditioning model to manipulate feature maps of a target CNN by scaling them up or down, negating them, shutting them off, selectively thresholding them, and more. Each feature map is modulated independently, giving the conditioning model an exponential (in the number of feature maps) number of ways to affect the feature representation.
    
    Rather than output $\hat{\gamma}_{i,c}$ directly, we output $\Delta\hat{\gamma}_{i,c}$, where:
\begin{equation}
	\hat{\gamma}_{i,c}=1+\Delta\hat{\gamma}_{i,c},
\end{equation}
since initially zero-centered $\hat{\gamma}_{i,c}$ can zero out CNN feature map activations and thus gradients. In our implementation, we opt to output $\Delta\hat{\gamma}_{i,c}$ rather than $\hat{\gamma}_{i,c}$, but for simplicity, in the rest of this paper, we will explain our method using $\hat{\gamma}_{i,c}$.

	\subsection{Model}
		\label{model}

        \begin{figure}[t]
            \centering
            \includegraphics[width=0.9\linewidth]{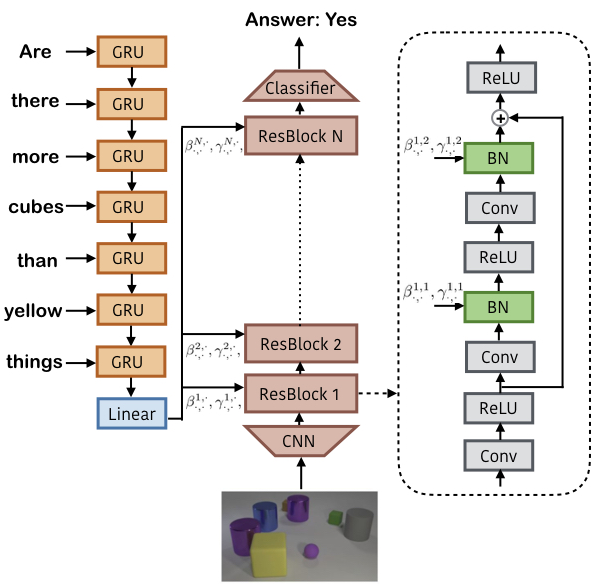}
	            \caption{\it{The linguistic pipeline (left), visual pipeline (middle), and CBN residual block architecture (right) of our model.}}
            \label{fig:model}
            \vskip -1em
        \end{figure}

		Our model consists of a linguistic pipeline and a visual pipeline as depicted in Figure~\ref{fig:model}. The linguistic pipeline processes a question $q$ using a Gated Recurrent Unit (GRU)~\cite{DBLP:journals/corr/ChungGCB14} with 4096 hidden units that takes in learned, 200-dimensional word embeddings. The final GRU hidden state is a question embedding $\bm{e}_q$. From this embedding, the model predicts the CBN parameters $(\bm{\gamma}^{m,n}_{i,\cdot}, \bm{\beta}^{m,n}_{i,\cdot})$ for the $n^{th}$ CBN layer of the $m^{th}$ residual block via linear projection with a trainable weight matrix $\bm{W}$ and bias vector $\bm{b}$:
        \begin{equation}
        (\bm{\gamma}^{m,n}_{i,\cdot}, \bm{\beta}^{m,n}_{i,\cdot}) = \bm{W}^{m,n}\bm{e}_q + \bm{b}^{m,n}
        \end{equation}

        The visual pipeline extracts $14\times14$ image features using the \textit{conv4} layer of a ResNet-101~\cite{DBLP:journals/corr/HeZRS15} pre-trained on ImageNet \cite{DBLP:journals/corr/RussakovskyDSKSMHKKBBF14}, as done in \cite{DBLP:journals/corr/JohnsonHMHLZG17} for CLEVR. Image features are processed by a $3\times3$ convolution followed by several --- 3 for our model --- CBN residual blocks with 128 feature maps, and a final classifier. The classifier consists of a $1\times1$ convolution to 512 feature maps, global max-pooling, and a two-layer MLP with 1024 hidden units that outputs a distribution over final answers.

        Each CBN residual block starts with a $1\times1$ convolution followed by two $3\times3$ convolutions with CBN as depicted in Figure~\ref{fig:model}. Drawing from~\cite{DBLP:journals/corr/SantoroRBMPBL17,DBLP:journals/corr/WattersTWPBZ17}, we concatenate coordinate feature maps indicating relative spatial position (scaled from $-1$ to $1$) to the image features, each residual block's input, and the classifier's input. We train our model end-to-end from scratch with Adam (learning rate $3e^{-4}$) \cite{DBLP:journals/corr/KingmaB14}, early stopping on the validation set, weight decay ($1e^{-5}$), batch size 64, and BN and ReLU throughout the visual pipeline, using only image-question-answer triplets from the training set.

\section{Experiments} \label{experiments}

	\subsection{CLEVR dataset}
    	CLEVR is a generated dataset of 700K (image, question, answer, program) tuples. Images contain 3D-rendered objects of various shapes, materials, colors, and sizes. Questions are multi-step and compositional in nature, as shown in Figure~\ref{fig:CLEVR}. They range from counting questions (\emph{"How many green objects have the same size as the green metallic block?"}) to comparison questions (\emph{"Are there fewer tiny yellow cylinders than yellow metal cubes?"}) and can be 40+ words long. Answers are each one word from a set of $28$ possible answers. Programs are an additional supervisory signal consisting of step-by-step instructions, such as \texttt{filter\_shape[cube]}, \texttt{relate[right]}, and \texttt{count}, on how to answer the question. Program labels are difficult to generate or come by for real world datasets. Our model avoids using this extra supervision, learning to reason effectively directly from linguistic and visual input.

	\subsection{Results}
        \begin{table*}[ht!]
        \centering
        
        \vspace{-1em}
        {\small
        \begin{tabular}{l|c|cccccc}
        \toprule
        {Model} & {\textbf{Overall}} & {Count} & {Exist} & \begin{tabular}{@{}c@{}}Compare \\ Numbers\end{tabular} & \begin{tabular}{@{}c@{}}Query \\ Attribute\end{tabular} & \begin{tabular}{@{}c@{}}Compare \\ Attribute\end{tabular}\\
        \midrule
        Human \cite{DBLP:journals/corr/JohnsonHMHLZG17}                   & 92.6 &86.7 &96.6 &86.5 &95.0 &96.0\\
        \midrule
        Q-type baseline \cite{DBLP:journals/corr/JohnsonHMHLZG17}         &41.8 &34.6 &50.2 &51.0 &36.0 &51.3\\
        LSTM \cite{DBLP:journals/corr/JohnsonHMHLZG17}                    &46.8 &41.7 &61.1 &69.8 &36.8 &51.8\\
        CNN+LSTM \cite{DBLP:journals/corr/JohnsonHMHLZG17}                &52.3 &43.7 &65.2 &67.1 &49.3 &53.0\\
        CNN+LSTM+SA \cite{DBLP:journals/corr/SantoroRBMPBL17}             &76.6 &64.4 &82.7 &77.4 &82.6 &75.4\\
        N2NMN* \cite{DBLP:journals/corr/HuARDS17}                         &83.7 &68.5 &85.7 &84.9 &90.0 &88.7\\
        PG+EE (9K prog.)* \cite{DBLP:journals/corr/JohnsonHMHLZG17}       &88.6 &79.7 &89.7 &79.1 &92.6 &96.0\\
        PG+EE (700K prog.)* \cite{DBLP:journals/corr/JohnsonHMHLZG17}     &96.9 &92.7 &97.1 &\bf{98.7} &98.1 &98.9\\
        CNN+LSTM+RN\textdagger \cite{DBLP:journals/corr/SantoroRBMPBL17}  &95.5 &90.1 &97.8 & 93.6 &97.9 &97.1\\
        \midrule
        CNN+GRU+CBN &\bf{97.6} &\bf{94.5} &\bf{99.2} &93.8 &\bf{99.2} &\bf{99.0}\\
        \bottomrule
        \end{tabular}
        }
        \caption{\it{\label{tab:results}
        CLEVR accuracy by baseline methods, competing methods, and our method (CBN). Methods denoted with (*) use extra supervisory information through program labels. Methods denoted with (\textdagger) use data augmentation and no pre-trained CNN.}}
        \end{table*}
		Our results on CLEVR are shown in Table~\ref{tab:results}. Our model achieves a new overall state-of-the-art, outperforming humans and previous, leading models, which often use additional program supervision. Notably, CBN outperforms Stacked Attention networks (CNN+LSTM+SA in \ref{tab:results}) by 21.0\%. Stacked Attention networks are highly effective for visual question answering with simpler questions~\cite{DBLP:journals/corr/YangHGDS15} and are the previously leading model for visual reasoning that does not build in reasoning, making them a relevant baseline for CBN. We note also that our model's pattern of performance more closely resembles that of humans than other models do. Strong performance ($<1\%$ error) in \texttt{exist} and \texttt{query\_attribute} categories is perhaps explained by our model's close resemblance to standard CNNs, which traditionally excel at these classification-type tasks. Our model also demonstrates strong performance on more complex categories such as \texttt{count} and \texttt{compare\_attribute}.
        
        Comparing numbers of objects gives our model more difficulty, understandably so; this question type requires more high-level reasoning steps --- querying attributes, counting, and comparing --- than other question type. The best model from \cite{DBLP:journals/corr/JohnsonHMHLZG17} beats our model here but is trained with extra supervision via 700K program labels. As shown in Table~\ref{tab:results}, the equivalent, more comparable model from \cite{DBLP:journals/corr/JohnsonHMHLZG17} which uses 9K program labels significantly underperforms our method in this category.

	\subsection{What does conditional batch norm learn?}

\begin{figure*}[th]
        \centering
        \includegraphics[width=0.9\linewidth]{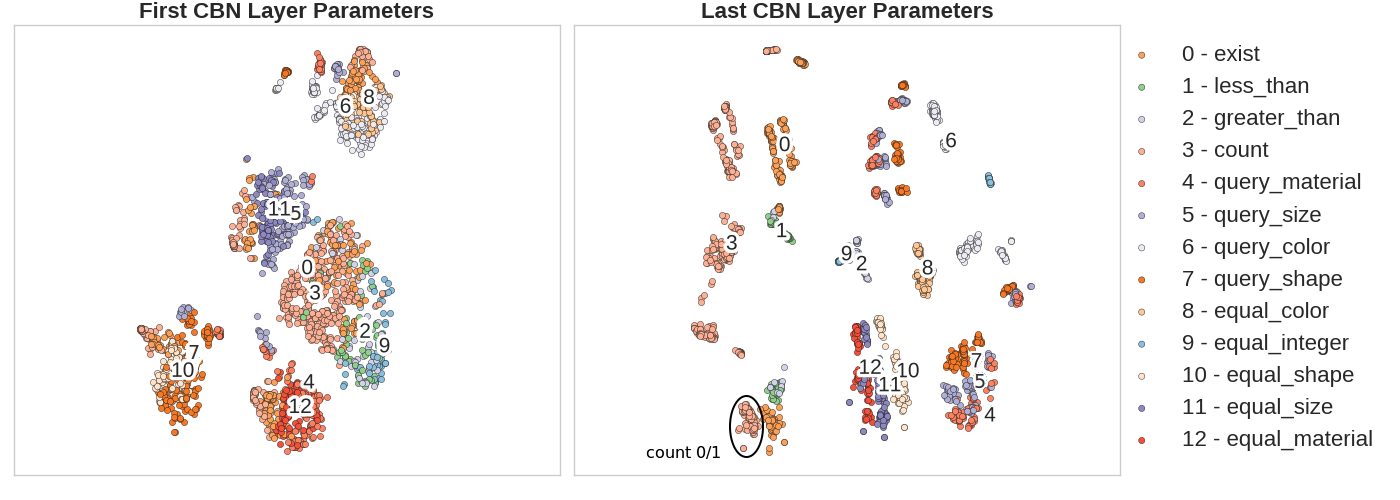}
        \caption{\it{t-SNE plots of  $\bm{\gamma}$, $\bm{\beta}$  of the first BN layer of the first residual block (left) and the last BN layer of the last residual block (right). CBN parameters are grouped by low-level reasoning functions for the first layer and by high-level reasoning functions for the last layer.}
        }
        \label{fig:tsne}
        \end{figure*}

    	To understand what our model learns, we use t-SNE~\cite{maaten2008visualizing} to visualize the CBN parameter vectors $(\bm{\gamma}, \bm{\beta})$, of 2,000 random validation points, modulating first and last CBN layers in our model, as shown in Figure~\ref{fig:tsne}. The $(\bm{\gamma}, \bm{\beta})$ parameters of the first and last CBN layers are grouped by the low-level and high-level reasoning functions necessary to answer CLEVR questions, respectively. For example, the CBN parameters for \texttt{equal\_color} and \texttt{query\_color} are close for the first layer but apart for the last layer, and the same is true for \texttt{equal\_shape} and \texttt{query\_shape}, \texttt{equal\_size} and \texttt{query\_size}, and \texttt{equal\_material} and \texttt{query\_material}. Conversely, \texttt{equal\_shape}, \texttt{equal\_size}, and \texttt{equal\_material} CBN parameters are grouped in the last layer but split in the first layer. Similar patterns emerge when visualizing residual block activations. Thus, we see that CBN learns a sort of function-based modularity, directly from language and image inputs and without an architectural prior on modularity. Simply with end-to-end training, our model learns to handle not only different types of questions differently, but also different types of question sub-parts differently, working from low-level to high-level processes as is the proper approach to answer CLEVR questions.

        Additionally, we observe that many points that break the previously mentioned clustering patterns do so in meaningful ways. For example, Figure~\ref{fig:tsne} shows that some \texttt{count} questions have last layer CBN parameters far from those of other \texttt{count} questions but close to those of \texttt{exist} questions. Closer examination reveals that these \texttt{count} questions have answers of either 0 or 1, making them similar to \texttt{exist} questions.
    
	\subsection{Error analysis}
		An analysis of our model's errors reveals that 94\% of its counting mistakes are off-by-one errors, indicating our model has learned underlying concepts behind counting, such as close relationships between close numbers.
        
\begin{figure}[t]
            \centering
            \includegraphics[width=1\linewidth]{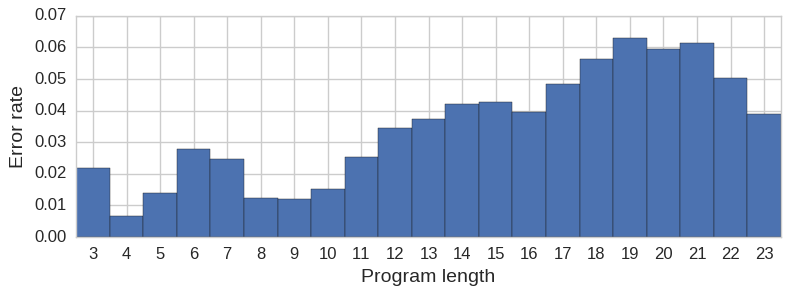}
            \caption{\it{Validation error rate by program length.}}
             \label{fig:errorrate}
            \vskip -1em
        \end{figure}

        As shown in Figure~\ref{fig:errorrate}, our CBN model struggles more on questions that require more steps, as indicated by the length of the corresponding CLEVR programs; error rates for questions requiring 10 or fewer steps are around $1.5\%$, while error rates for questions requiring 17 or more steps are around $5.5\%$, more than three times higher.

\begin{figure}[ht]
        \center
            \includegraphics[width=0.7\linewidth]{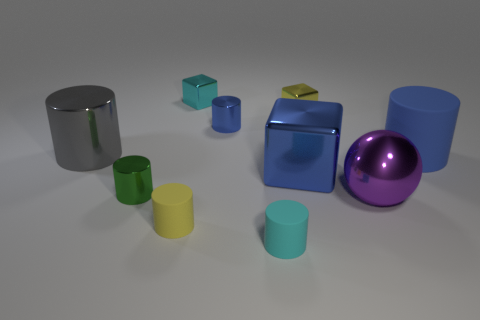}

        \vskip 1em
        \scriptsize
        \begin{tabular}{|l|l|}
        \hline
        Question & Answer\\
        \hline
        How many yellow things are there? & 2\\
        How many cyan things are there? & 2\\
        Are there as many yellow things as cyan things? & \bf{No}\\
        Are there more yellow things than cyan things? & No\\
        Are there fewer yellow things than cyan things? & No\\
        \hline
        \end{tabular}
        \caption{\it{An interesting failure example where our model counts correctly but compares counts erroneously. Its third answer is incorrect and inconsistent with its other answers.}}
        \label{fig:failure}
        \vskip -1em
        \end{figure}

    	Furthermore, the model sometimes makes curious reasoning mistakes a human would not. In Figure~\ref{fig:failure}, we show an example where our model correctly counts two cyan objects and two yellow objects but simultaneously does not answer that there are the same number of cyan and yellow objects. In fact, it does not answer that the number of cyan blocks is more, less, \textit{or} equal to the number of yellow blocks. These errors could be prevented by directly minimizing logical inconsistency, which is an interesting avenue for future work orthogonal to our approach.
        
        These types of mistakes in a state-of-the-art visual reasoning model suggest that more work is needed to truly achieve human-like reasoning and logical consistency. We view CLEVR as a curriculum of tasks and believe that the key to the most meaningful and advanced reasoning lies in tackling these last few percentage points of error.

\section{Related Work} \label{relatedwork}

	One leading approach for visual reasoning is the Program Generator + Execution Engine model from \cite{DBLP:journals/corr/JohnsonHMHLZG17}. This approach consists of a sequence-to-sequence “Program Generator” (PG), which takes in a question and outputs a sequence corresponding to a tree of composable Neural Modules, each of which is a two-layer residual block similar to ours. This tree of Neural Modules is assembled to form the Execution Engine (EE) that then predicts an answer from the image. The PG+EE model uses a strong prior by training with program labels and explicitly modeling the compositional nature of reasoning. Our approach learns to reason directly from textual input without using additional cues or a specialized architecture.

	This modular approach is part of a recent line of work in Neural Module Networks \cite{DBLP:journals/corr/HuARDS17,DBLP:journals/corr/AndreasRDK15,DBLP:journals/corr/AndreasRDK16}. Of these, End-to-End Module Networks (N2NMN) \cite{DBLP:journals/corr/HuARDS17} also tackle visual reasoning but do not perform as well as other approaches. These methods also use strong priors by modeling the compositionality of reasoning, using program-level supervision, and building per-module, hand-crafted neural architectures for specific functions.\\

	Relation Networks (RNs) from \cite{DBLP:journals/corr/SantoroRBMPBL17} are another leading approach for visual reasoning. RNs use an MLP to carry out pairwise comparisons over each location of extracted convolutional features over an image, including LSTM-extracted question features as input to this MLP. RNs then element-wise sum over the resulting comparison vectors to form another vector from which a final classifier predicts the answer. This approach is end-to-end differentiable and trainable from scratch to high performance, as we show in Table~\ref{tab:results}. Our approach lifts the explicitly relational aspect of this model, freeing our approach from the use of a comparison-based prior, as well as the scaling difficulties of pairwise comparisons over spatial locations. 

	CBN itself has its own line of work. The results of \cite{DBLP:journals/corr/DumoulinSK16,DBLP:journals/corr/GhiasiLKDS17} show that the closely related Conditional Instance Normalization is able to successfully modulate a convolutional style-transfer network to quickly and scalably render an image in a huge variety of different styles, simply by learning to output a different set of BN parameters based on target style. For visual question answering, answering general questions often of natural images, de Vries \etal~\cite{modulating_vision} show that CBN performs highly on real-world VQA and GuessWhat?! datasets, demonstrating CBN's effectiveness beyond the simpler CLEVR images. Their architecture conditions 50 BN layers of a pre-trained ResNet. We show that a few layers of CBN after a ResNet can also be highly effective, even for complex problems. We also show how CBN models can learn to carry out multi-step processes and reason in a structured way --- from low-level to high-level.

	Additionally, CBN is essentially a post-BN, feature-wise affine conditioning, with BN's trainable scalars turned off. Thus, there are many interesting connections with other conditioning methods. A common approach, used for example in Conditional DCGANs~\cite{DBLP:journals/corr/RadfordMC15}, is to concatenate constant feature maps of conditioning information to the input of convolutional layers, which amounts to adding a post-convolutional, feature-wise conditional bias. Other approaches, such as LSTMs~\cite{LSTM} and Hierarchical Mixtures of Experts~\cite{HME}, gate an input's features as a function of that same input (rather than a separate, conditioning input), which amounts to a feature-wise, conditional scaling, restricted to between 0 and 1. CBN consists of both scaling and shifting, each unrestricted, giving it more capacity than many of these related approaches. We leave exploring these connections more in-depth for future work.

\section{Conclusion} \label{conclusions}

	With a simple and general model based on CBN, we show it is possible to achieve state-of-the-art visual reasoning on CLEVR without explicitly incorporating reasoning priors. We show that our model learns an underlying structure required to answer CLEVR questions by finding clusters in the CBN parameters of our model; earlier parameters are grouped by low-level reasoning functions while later parameters are grouped by high-level reasoning functions. Simply by manipulating feature maps with CBN, a RNN can effectively use language to influence a CNN to carry out diverse and multi-step reasoning tasks over an image. It is unclear whether CBN is the most effective general way to use conditioning information for visual reasoning or other tasks, as well as what precisely about CBN is so effective. Other approaches \cite{DBLP:journals/corr/RadfordMC15, LSTM, HME, DBLP:journals/corr/OordDZSVGKSK16,van2016conditional,DBLP:journals/corr/ReedOKCWBF17,reed2016generating} employ a similar, repetitive conditioning, so perhaps there is an underlying principle that explains the success of these approaches. Regardless, we believe that CBN is a general and powerful technique for multi-modal and conditional tasks, especially where more complex structure is involved.

  \section{Acknowledgements}

    We would like to thank the developers of PyTorch (\url{http://pytorch.org/}) for their elegant deep learning framework. Also, our implementation was based off the open-source code from \cite{DBLP:journals/corr/JohnsonHMHLZG17}. We thank Mohammad Pezeshki, Dzmitry Bahdanau, Yoshua Bengio, Nando de Freitas, Joelle Pineau, Olivier Pietquin, J\'er\'emie Mary, Chin-Wei Huang, Layla Asri, and Max Smith for helpful feedback and discussions, as well as Justin Johnson for CLEVR test set evaluations. We thank NVIDIA for donating a DGX-1 computer used in this work. We also acknowledge FRQNT through the CHIST-ERA IGLU project and CPER Nord-Pas de Calais, Coll\`ege Doctoral Lille Nord de France and FEDER DATA Advanced data science and technologies 2015-2020 for funding our research.
    
 
  \eightpt
  \bibliographystyle{IEEEtran}
  \bibliography{mybib}

\end{document}